\documentclass[11pt,letterpaper]{article}
\usepackage[letterpaper]{geometry}
\usepackage{emnlp2018}
\usepackage{times}
\usepackage{latexsym}
\usepackage{amsmath}
\usepackage{multirow}
\usepackage{url}
\usepackage{amsthm,amsmath,amssymb,graphicx}
\usepackage{algorithm2e}
\usepackage{algpseudocode}

\makeatletter
\newcommand{\@emptybiblabel}[1]{}
\makeatother

\usepackage{amsfonts}
\usepackage{graphicx}
\graphicspath{ {.} }
\usepackage{tabularx}

\usepackage{booktabs} 
\usepackage{caption}
\usepackage{subcaption}

\usepackage{enumitem}
\setenumerate{leftmargin=5.5mm,itemsep=0pt,topsep=3pt}

\usepackage{tikz}
\usetikzlibrary{positioning}
\usetikzlibrary{shapes.geometric, arrows}
\tikzstyle{startstop} = [rectangle, rounded corners, minimum width=1cm, minimum height=0.5cm,text centered, draw=black]
\tikzstyle{io} = [trapezium, trapezium left angle=70, trapezium right angle=110, minimum width=1cm, minimum height=0.5cm, text centered, draw=black]
\tikzstyle{process} = [rectangle, minimum width=1cm, minimum height=0.5cm, text centered, draw=black]
\tikzstyle{decision} = [diamond, inner sep=0.05cm, aspect=2, text centered, draw=black]
\tikzstyle{arrow} = [thick,->,>=stealth]

\newcommand{\eat}[1]{\ignorespaces}

\newcommand{\commentout}[1]{}

\newcommand{\gleu}{\textrm{GLEU$^+$}}

\newcommand{\ffive}{\textrm{$F_{0.5}$ }}

\colorlet{gold}{green!10!orange!90!}

\newcommand{\ignore}[1]{}

\aclfinalcopy
\title{Weakly Supervised Grammatical Error Correction using \\ Iterative Decoding}

\author{Jared Lichtarge, Christopher Alberti, Shankar Kumar, Noam Shazeer, Niki Parmar \\
  Google AI \\
  \{lichtarge,chrisalberti,shankarkumar,noam,nikip\}@google,com \\ }
\date{}

\begin{document}

\maketitle
\begin{abstract}
  We describe an approach to Grammatical Error Correction (GEC)
  that is effective at making use of models trained on large amounts of
  weakly supervised bitext. We train the \textit{Transformer}
  sequence-to-sequence model on 4B tokens of Wikipedia revisions and
  employ an iterative decoding strategy that is tailored to
  the loosely-supervised nature of the Wikipedia training corpus.
  Finetuning on the Lang-8 corpus and ensembling yields an
  \ffive of 58.3 on the CoNLL'14 benchmark and a GLEU of 62.4 on JFLEG.
  The combination of weakly supervised training and iterative decoding obtains an \ffive
  of 48.2 on CoNLL'14 even without using any labeled GEC data.
\end{abstract}

\section{Introduction}
\label{sec:intro}

Much progress in the Grammatical Error Correction
(GEC) task can be credited to approaching the problem as a translation
task~\cite{brockett2006correcting} from an ungrammatical source language to a grammatical target language. This strict analogy to translation imparts an unnecessary all-at-once constraint.
We hypothesize that GEC can be more accurately characterized as a multi-pass iterative process, in which progress is made incrementally through the accumulation of minor corrections (Table~\ref{tab:decode}).
We address the relative scarcity of publicly available GEC training data by leveraging the entirety of English language Wikipedia revision histories\footnote{\url{https://dumps.wikimedia.org/enwiki/latest/}}, a large corpus that is weakly supervised for GEC because it only occasionally contains grammatical error corrections and is not human curated specifically for GEC.

\begin{table}[]
  \footnotesize
  \centering
  \begin{tabular}{p{1.3cm}|p{5cm}}
    \toprule
    Original & this is nto the pizzza that i ordering \\
    1st & this is \textbf{not} the \textbf{pizza} that \textbf{I} ordering \\
    2nd & \textbf{T}his is not the pizza that I ordering \\
    3nd & This is not the pizza that I \textbf{ordered} \\
    4th & This is not the pizza that I ordered\textbf{.} \\
    Final & This is not the pizza that I ordered. \\
    \bottomrule
  \end{tabular}
  \caption{Iterative decoding on a sample sentence. }
  \label{tab:decode}
\end{table}

In this work, we present an iterative decoding algorithm that allows for incremental corrections. While prior work~\cite{dahlmeier2012iterative} explored a similar algorithm to progressively expand the search space for GEC using a phrase-based machine translation approach, we demonstrate the effectiveness of this approach as a means of domain transfer for models trained exclusively on noisy out-of-domain data.
%
%

We apply iterative decoding to a {\it Transformer} model~\cite{vaswani2017attention} trained on minimally-filtered Wikipedia revisions, and show the model is already useful for GEC. With finetuning on Lang-8, our approach achieves the best reported single model result on the CoNLL'14 GEC task, and by ensembling four models, we obtain the state-of-the-art.

\section{Pretraining Data}
\label{sec:datagen}
Wikipedia is a publicly available, online encyclopedia for which all
content is communally created and curated. We use the revision histories of Wikipedia pages as training data for GEC. Unlike the WikEd corpus for GEC~\cite{grundkiewicz2014wiked}, our extracted corpus does not include any heuristic grammar-specific filtration beyond simple text extraction and is two orders of magnitude larger than Lang-8~\cite{mizumoto2011mining}, the largest publicly available corpus curated for GEC (Table~\ref{tab:tokens}). Section~\ref{sec:experiments} describes our data generation method.

\begin{table}[h]
  \footnotesize
  \centering
  \begin{tabular}{c|c|c}
    \toprule
    Corpus & Num. of sentences & Num. of words \\
    \midrule
    Wikipedia revisions & 170M  & 4.1B  \\
    Lang-8 & 1.9M & 25.0M \\
    WikEd & 12M & 292 M \\
    \bottomrule
  \end{tabular}
  \caption{Statistics computed over training sets for GEC.}
  \label{tab:tokens}
\end{table}

In Table~\ref{tab:wikiexamples}, we show
representative examples of the extracted source-target pairs,
including some artificial errors. While some of the edits
are grammatical error corrections, the vast majority are not.

\begin{table}[]
  \footnotesize
  \centering
  \begin{tabular}{p{1.1cm}|p{5cm}}
    \toprule
    Original & Artilleryin 1941 \textbf{and} was medically discharged \\
    Target   & Artilleryin 1941 \textbf{he} was \textbf{later} medically
    discharged \textbf{with} \\
    \midrule
    Original & \textbf{Wolfpac} has their \textbf{evry}
    own \textbf{i}nternet radio show \\
    Target   & \textbf{WOLFPAC} has their \textbf{very}
    own \textbf{I}nternet radio show \\
    \midrule
    Original & League called ONEFA. TEXTBFhe University \textbf{is also a site for the third} \\
    Target & League called ONEFA. The University \textbf{also hosts the third Spanish} \\
    \bottomrule
  \end{tabular}
  \caption{Example source-target pairs from the Wikipedia
    dataset used for pretraining models. }
  \label{tab:wikiexamples}
\end{table}

\section{Decoding}
\label{sec:decode}
Our iterative decoding algorithm is presented in
Algorithm~\ref{algo:decode}. Unlike supervised
bitext such as CoNLL, our Wikipedia-derived bitext typically
contains fewer edits. Thus a model trained on Wikipedia
learns to make very few edits in a single decoding pass. Iterative
decoding alleviates this problem by applying a sequence of rewrites
starting from the grammatically incorrect sentence ($I$), making incremental
improvements until it cannot find any more edits to make. In each iteration,
the algorithm performs a conventional beam search
but is only allowed to output a rewrite for which it has
high confidence. The best non-identity decoded target
sentence is output only if its cost is less than the cost of the
identity translation times a predetermined threshold.

\begin{figure}[hb]
	\centering
	\includegraphics[height=2in]{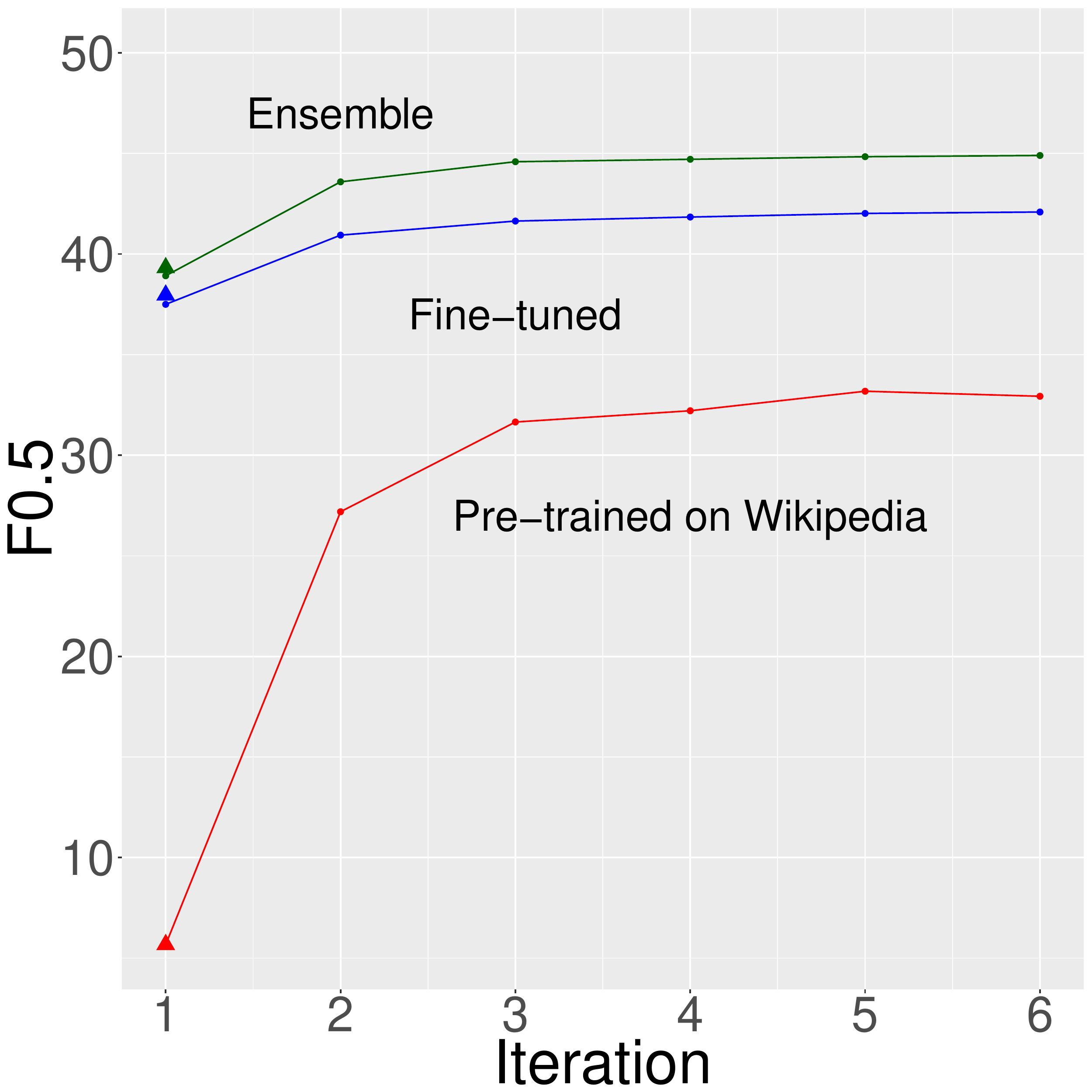}
        \caption{\ffive with iterative decoding on the CoNLL dev set. Triangles indicate performance with single-shot decoding. Each point for the pre-trained/fine-tuned settings is an average performance across 4 models.}
	\label{fig:iterative}
\end{figure}

Applied to the models trained exclusively on out-of-domain
Wikipedia data, iterative decoding mediates domain transfer by allowing the
accumulation of incremental changes, as would be more typical of Wikipedia,
rather than requiring a single-shot fix, as is the format of curated
GEC data. Using incremental edits produces a significant improvement
in performance over single-shot decoding, revealing that the pre-trained models, which would have otherwise
appeared useless, may already be useful for GEC by themselves
(Figure~\ref{fig:iterative}). The improvements from iterative decoding on
finetuned models are not as dramatic, but still substantial.

\RestyleAlgo{boxruled}

\begin{algorithm}
\small
\KwData{$I$, $\text{beam}$, $\text{threshold}$, $\text{MAXITER}$}
\KwResult{$\hat{T}$}
\For{$i \in \{1,2,...,\text{MAXITER}\}$} {
  $\text{Nbestlist} = \text{Decode}(I, \text{beam})$ \\
  $C_{\text{Identity}} = +\infty$ \\
  $C_{\text{Non-Identity}} = +\infty$ \\
  $H_{\text{Non-Identity}} = NULL $ \\
  \For{$H \in \text{Nbestlist}$} {
    \uIf{$H = I$} {
      $C_{\text{Identity}} = \text{Cost}(H)$; \\
    }
    \uElseIf {\text{Cost}(H) $<$ $C_\text{Non-Identity}$} {
      $C_{\text{Non-Identity}} = \text{Cost}(H)$ \\
      $H_{\text{Non-Identity}} = H$
    }
  }
  \Comment{Rewrite if non-identity cost $<$ identity cost}\\
  \eIf{$C_{\text{Non-Identity}} / C_{\text{Identity}} < \text{threshold}$} {
    $\hat{T} = H_{\text{Non-Identity}}$ \Comment{Output rewrite.}
  }{
    $\hat{T} = I$ \Comment{Output identity.}
  }
  $I = \hat{T}$ \Comment{Input for next iteration.}
}
\caption{Iterative Decoding}
\label{algo:decode}
\end{algorithm}

In Table~\ref{tab:decode}, we show an example of iterative decoding in
action. The model continues to refine the input until it reaches a
sentence that does not require any edits. We generally see fewer edits
being applied as the model gets closer to the final result.

\section{Model}
\label{sec:model}
In this work, we use the \textit{Transformer} sequence-to-sequence model~\cite{vaswani2017attention}, using the \textit{Tensor2Tensor} opensource
implementation.\footnote{\url{https://github.com/tensorflow/tensor2tensor}} We use 6 layers for both the encoder and the decoder, 8 attention heads, a dictionary of 32k word pieces~\cite{schuster2012wordpiece}, embedding size $d_{model} = 1024$, a position-wise feed forward network at every layer of inner size $d_{ff} = 4096$, and Adafactor as optimizer with inverse squared root decay~\cite{shazeer2018adafactor}.\footnote{We used the ``transformer\_clean\_big\_tpu'' setting.}

\section{Experiments}
\label{sec:experiments}

\begin{table*}[ht]
  \footnotesize
  \centering
  \begin{tabular}{p{1.3cm}|p{13cm}}
    \toprule
    Original & Recently, a \textbf{new coming} surveillance technology called radio-frequency identification \textbf{which is RFID for short} has caused heated discussions on whether it should be used to track people. \\
    Pretrained & Recently, a surveillance technology called radio frequency identification \textbf{(RFID)} has caused heated discussions on whether it should be used to track people. \\
    Finetuned & Recently, a \textbf{new} surveillance technology called radio-frequency identification\textbf{, which is RFID for short,} has caused heated discussions on whether it should be used to track people. \\
    \midrule
    Original & \textbf{Then we can see that t}he rising life \textbf{expectancies} can also be viewed as a challenge for \textbf{us} to face. \\
    Pretrained & \textbf{T}he rising life \textbf{expectancy} can also be viewed as a challenge for \textbf{people} to face. \\
    Finetuned & \textbf{Then we can see that t}he rising life \textbf{expectancy} can also be viewed as a challenge for \textbf{us} to face. \\
    \bottomrule
  \end{tabular}
  \caption{Corrections from the pretrained/finetuned-ensemble models
    on example sentences from the CoNLL'14 dev set. }
  \label{tab:examples}
\end{table*}

Starting with the raw XML of the \textit{Wikipedia} revision history dump,
we extract individual pages, each containing snapshots in chronological order. We extract
the inline text and remove the non-text elements within. We throw out pages larger than 64Mb. For remaining pages, we logarithmically downsample pairs of consecutive snapshots,
admitting only $\log_{1.5}(X)$ pairs for a total of $X$ snapshots.\footnote{This prevents larger pages with more snapshots from overwhelming smaller pages, and reduces the total amount of data 20-fold.} Each remaining
pair of consecutive snapshots forms a source/target pair.

Our goal is to train a single model that can perform both spelling and grammar correction. We therefore introduce spelling errors on the source side at a rate of $0.003$ per
character, using deletion, insertion, replacement, and transposition of adjacent characters. We then align the texts from consecutive snapshots and extract sequences between matching segments with a maximum length of 256 word-pieces.\footnote{An alternative approach would have been to extract full sentences, but we decided against introducing the complexity of a model for identifying sentence boundaries.} Examples with identical source and target sequences are downsampled by 99\% to achieve 3.8\% identical examples in the final data.

We experimented with data filtering by discarding examples where
source and target were further than a maximum edit distance apart, by varying the max page size cutoff, and trying different rates of downsampling consecutive pages. Models trained on the augmented data did not obtain substantially different
performance. We did however observe performance improvements when we
ensembled together models trained on datasets with different filtering
settings.

We train the {\it Transformer} model on Wikipedia
revisions for 5 epochs with a batch size of approximately 64,000 word pieces. During this pre-training,
we set the learning rate to 0.1 for the first 10,000 steps, then
decrease it proportionally to the inverse square root of the number
of steps after that. We average the weights of the model over 8 checkpoints spanning the final 1.5 epochs of training.

We then finetune our models on Lang-8 for 50 epochs, linearly
increasing the learning rate from 0 to $3 \cdot 10^{-5}$ over the
first 20,000 steps and keeping the learning rate constant for the
remaining steps. We stop the fine-tuning before the models start to
overfit on a development set drawn from Lang-8.

At evaluation time, we run iterative decoding using a beam size of 4.
Finally, we apply a small set of regular expressions
to match the tokenization to that of the dataset. Our
ensemble models are obtained by decoding with 4 identical {\it Transformers}
pretrained and finetuned separately. At each step of decoding, we average the logits from the
4 models.

Following~\cite{grundkiewicz2018near,junczys2018approaching}, we
preprocess JFLEG development and test sets with a spell-checking component but do not apply spelling
correction to CoNLL sets. For CoNLL sets, we pick the iterative
decoding threshold and number of iterations
on a subset of the CoNLL'14 training set, sampled
to have the same ratio of modified to unmodified sentences as the
CoNLL'14 dev set. For JFLEG, we pick the best decoding threshold on the
JFLEG dev set.  We report performance of our models by measuring
\ffive with the $M^2$ scorer (\newcite{dahlmeier2012better}) on the
CoNLL'14 dev and test sets, and the GLEU+ metric~\cite{napoles2016gleu} on the JFLEG dev and test sets.

The results of our method are shown in Table~\ref{tab:results}. On both CoNLL'14 and JFLEG, we achieve state-of-the-art
for both single models and ensembles. In all cases, iterative decoding substantially outperforms single shot decoding.

\begin{table*}[ht]
    \centering
    \footnotesize
    \begin{tabular}{c|cc|rrrr|rr}
    \toprule
        & & & \multicolumn{4}{c|}{CoNLL14} & \multicolumn{2}{c}{JFLEG} \\
        & & & dev & \multicolumn{3}{c|}{test} & dev & test \\
        & & & \ffive & Precision & Recall & \ffive & \multicolumn{2}{c}{\gleu} \\
    \midrule
    (1) & \multicolumn{2}{c|}{MLConv$_{embed}$} &                   
                                            & 60.9 & 23.7 & 46.4  
                                            & 47.7 & 51.3 \\       
        & \multicolumn{2}{c|}{MLConv$_{embed}$ (4 ensemble) +EO +LM +SpellCheck} &              
                                            & 65.5 & 33.1 & 54.8  
                                            & 52.5 & 57.5 \\       
    \midrule
    (2) & \multicolumn{2}{c|}{Transformer (single)}    &            
                                            & & & 53.0     
                                            & & 57.9 \\  
        & \multicolumn{2}{c|}{Transformer (4 ensemble)}  & 41.5                
                                            & 63.0 & 38.9 & 56.1  
                                            & & 58.5 \\           
        & \multicolumn{2}{c|}{Transformer (4 ensemble) +LM}          & 42.9                
                                            & 61.9 & 40.2 & 55.8  
                                            & & 59.9 \\          
    \midrule
    (3) & \multicolumn{2}{c|}{Hybrid PBMT +NMT +LM}    &                        
                                            & 66.8 & 34.5 & 56.3  
                                            & & 61.5 \\    
    \midrule
    This work & Model & Decoding Type & & & & & &  \\
    \midrule
    & Transformer (single, pretrained) & single-shot & 5.7
                                            & 63.0 & 7.2 & 24.6
                                            & 45.4 & 50.4    \\
    & Transformer (single, pretrained) & iterative & 33.2
                                            & 56.8 & 30.3 & 48.2
    & 51.1 & 56.1    \\
    & Transformer (single, finetuned) & single-shot & 38.0
                                            & 64.3 & 29.7 & 52.2
                                            & 51.3 & 56.6 \\
    & Transformer (single, finetuned) & iterative & 42.9
                                            & 62.2 & 37.8 & 54.9
    & 54.2 & 59.3  \\
    & Transformer (4 ensemble, finetuned) & single-shot & 39.3
                                            & 67.9 & 31.6 & 55.2
                                            & 52.6 & 57.9 \\
    & Transformer (4 ensemble, finetuned) & iterative  & \textbf{45.0}
                                            & 67.5 & 37.8 & \textbf{58.3}
                                            & 56.8 & \textbf{62.4} \\
    \bottomrule
    \end{tabular}
    \caption{Comparison of our model with recent state-of-the-art models on the CoNLL'14
      and JFLEG datsets. All single model results are averages of 4 models.
      (1): \newcite{chollampatt2018multilayer}, (2): \newcite{junczys2018approaching},
      (3): \newcite{grundkiewicz2018near}.
    }
    \label{tab:results}
\end{table*}

\section{Error Analysis}\label{sec:eval_analysis}

In Table~\ref{tab:examples}, we list example corrections proposed by
the model pretrained on Wikipedia revisions and by the ensemble model
finetuned on Lang-8. The changes proposed by the pretrained model
often appear to be improvements to the original sentence, but fall outside the scope of GEC.
Models finetuned on Lang-8 learn to make more conservative
corrections.

The finetuning on Lang-8 can be viewed as a domain adaptation
technique that shifts the pretrained model from the Wikipedia domain
to the GEC domain. On Wikipedia, it is common to see
substantial edits that make the text more concise and readable,
e.g. replacing ``which is RFID for short'' with ``(RFID)'', or
removing less important clauses like ``Then we can see that''. But
these are not appropriate for GEC as they are editorial style fixes
rather than grammatical fixes.

\section{Related Work}
\label{sec:related}

Progress in GEC has accelerated rapidly since
the CoNLL'14 Shared Task~\cite{ng2014conll}. \newcite{rozovskaya2016grammatical}
combined a Phrase Based Machine Translation (PBMT)
model trained on the Lang-8 dataset \cite{mizumoto2011mining} with
error specific classifiers. \newcite{junczys2016phrase} combined a PBMT model with bitext features
and a larger language model. The first Neural Machine Translation (NMT) model to reach
the state of the art on CoNLL'14~\cite{chollampatt2018multilayer} used an
ensemble of four convolutional sequence-to-sequence models followed by
rescoring. The current state of the art (\ffive of 56.25 on ConLL '14) was achieved by
\newcite{grundkiewicz2018near} with a hybrid PBMT-NMT system. A neural-only result with an \ffive of 56.1 on CoNLL '14
was reported by \newcite{junczys2018approaching}
using an ensemble of neural {\it Transformer} models
\cite{vaswani2017attention}, where the decoder side of each model is pretrained as a language model. Our approach can be viewed as a direct
extension of this last work, where our novel contributions include iterative decoding and the pretraining on a large amount of Wikipedia edits, instead of pretraining only the decoder as a language model. While pretraining on out-of-domain data has been employed previously for neural machine translation~\cite{luong2015pretrain}, it has not been presented in GEC thus far.

\section{Discussion}
\label{sec:discussion}

We presented a neural {\it Transformer} model that obtains state-of-the-art
results on CoNLL'14 and JFLEG tasks\footnote{Using non-public sentences crawled from \url{Lang-8.com},~\newcite{ge2018} recently obtained an \ffive of $61.34$ on CoNLL'14 and a GLEU of 62.4 on JFLEG.}. Our contributions are twofold: we couple the use of publicly
available Wikipedia revisions at much larger scale than previously
reported for GEC, with an iterative decoding strategy that
is especially useful when using models
trained on noisy bitext such as Wikipedia. Training on Wikipedia revisions alone
gives an \ffive of 48.2 on the CoNLL'14 task without relying on
human curated GEC data or non-parallel data. We also show that
a model trained using Wikipedia revisions can yield extra gains from
finetuning using the Lang-8 corpus and ensembling. We expect our work to spur interest
in methods for using noisy parallel data to improve NLP tasks.

\bibliographystyle{acl_natbib_nourl}
\bibliography{tgec}

\begin{thebibliography}{17}
\expandafter\ifx\csname natexlab\endcsname\relax\def\natexlab#1{#1}\fi

\bibitem[{Brockett et~al.(2006)Brockett, Dolan, and
  Gamon}]{brockett2006correcting}
Chris Brockett, William~B Dolan, and Michael Gamon. 2006.
\newblock Correcting {ESL} errors using phrasal {SMT} techniques.
\newblock In \emph{Proceedings of the 21st International Conference on
  Computational Linguistics and the 44th annual meeting of the Association for
  Computational Linguistics}, pages 249--256. Association for Computational
  Linguistics.

\bibitem[{Chollampatt and Ng(2018)}]{chollampatt2018multilayer}
Shamil Chollampatt and Hwee~Tou Ng. 2018.
\newblock A multilayer convolutional encoder-decoder neural network for
  grammatical error correction.
\newblock {arXiv}:1801.08831.

\bibitem[{Dahlmeier and Ng(2012{\natexlab{a}})}]{dahlmeier2012iterative}
Daniel Dahlmeier and Hwee~Tou Ng. 2012{\natexlab{a}}.
\newblock A beamsearch decoder for grammatical error correction.
\newblock In \emph{Joint Conference on Empirical Methods in Natural Language
  Processing and Computational Natural Language Learning}.

\bibitem[{Dahlmeier and Ng(2012{\natexlab{b}})}]{dahlmeier2012better}
Daniel Dahlmeier and Hwee~Tou Ng. 2012{\natexlab{b}}.
\newblock Better evaluation for grammatical error correction.
\newblock In \emph{Proc. of NAACL}.

\bibitem[{Grundkiewicz and Junczys-Dowmunt(2014)}]{grundkiewicz2014wiked}
Roman Grundkiewicz and Marcin Junczys-Dowmunt. 2014.
\newblock The wiked error corpus: A corpus of corrective wikipedia edits and
  its application to grammatical error correction.
\newblock In \emph{International Conference on Natural Language Processing},
  pages 478--490. Springer.

\bibitem[{Grundkiewicz and Junczys-Dowmunt(2018)}]{grundkiewicz2018near}
Roman Grundkiewicz and Marcin Junczys-Dowmunt. 2018.
\newblock Near human-level performance in grammatical error correction with
  hybrid machine translation.
\newblock \emph{{arXiv}:1804.05945}.

\bibitem[{Junczys-Dowmunt and Grundkiewicz(2016)}]{junczys2016phrase}
Marcin Junczys-Dowmunt and Roman Grundkiewicz. 2016.
\newblock Phrase-based machine translation is state-of-the-art for automatic
  grammatical error correction.
\newblock In \emph{Proc. of EMNLP}.

\bibitem[{Junczys-Dowmunt et~al.(2018)Junczys-Dowmunt, Grundkiewicz, Guha, and
  Heafield}]{junczys2018approaching}
Marcin Junczys-Dowmunt, Roman Grundkiewicz, Shubha Guha, and Kenneth Heafield.
  2018.
\newblock Approaching neural grammatical error correction as a low-resource
  machine translation task.
\newblock \emph{{arXiv}:1804.05940}.

\bibitem[{Luong and Manning(2015)}]{luong2015pretrain}
Minh-Thang Luong and Christopher~D. Manning. 2015.
\newblock Stanford neural machine translation systems for spoken language
  domain.
\newblock In \emph{International Workshop on Spoken Language Translation}.

\bibitem[{Mizumoto et~al.(2011)Mizumoto, Komachi, Nagata, and
  Matsumoto}]{mizumoto2011mining}
Tomoya Mizumoto, Mamoru Komachi, Masaaki Nagata, and Yuji Matsumoto. 2011.
\newblock Mining revision log of language learning {SNS} for automated japanese
  error correction of second language learners.
\newblock In \emph{Proceedings of 5th International Joint Conference on Natural
  Language Processing}, pages 147--155.

\bibitem[{Napoles et~al.(2016)Napoles, Sakaguchi, Post, and
  Tetreault}]{napoles2016gleu}
Courtney Napoles, Keisuke Sakaguchi, Matt Post, and Joel Tetreault. 2016.
\newblock {GLEU} without tuning.
\newblock {arXiv}:1605.02592.

\bibitem[{Ng et~al.(2014)Ng, Wu, Briscoe, Hadiwinoto, Susanto, and
  Bryant}]{ng2014conll}
Hwee~Tou Ng, Siew~Mei Wu, Ted Briscoe, Christian Hadiwinoto, Raymond~Hendy
  Susanto, and Christopher Bryant. 2014.
\newblock The {CoNLL-2014} shared task on grammatical error correction.
\newblock In \emph{CoNLL Shared Task}, pages 1--14.

\bibitem[{Rozovskaya and Roth(2016)}]{rozovskaya2016grammatical}
Alla Rozovskaya and Dan Roth. 2016.
\newblock Grammatical error correction: Machine translation and classifiers.
\newblock In \emph{Proc. of ACL}.

\bibitem[{Schuster and Nakajima(2012)}]{schuster2012wordpiece}
Michael Schuster and Kaisuke Nakajima. 2012.
\newblock Japanese and korean voice search.
\newblock In \emph{Proceedings of the IEEE Conference on Acoustics, Speech and
  Signal Processing}.

\bibitem[{Shazeer and Stern(2018)}]{shazeer2018adafactor}
Noam Shazeer and Mitchell Stern. 2018.
\newblock Adafactor: Adaptive learning rates with sublinear memory cost.
\newblock \emph{{arXiv}:1804.04235}.

\bibitem[{Tao et~al.(2018)Tao, Wei, and Zhou}]{ge2018}
Ge~Tao, Furu Wei, and Ming Zhou. 2018.
\newblock Reaching human-level performance in automatic grammar error
  correction: An empirical study.
\newblock {arXiv}:1807.01270.

\bibitem[{Vaswani et~al.(2017)Vaswani, Shazeer, Parmar, Uszkoreit, Jones,
  Gomez, Kaiser, and Polosukhin}]{vaswani2017attention}
Ashish Vaswani, Noam Shazeer, Niki Parmar, Jakob Uszkoreit, Llion Jones,
  Aidan~N Gomez, {\L}ukasz Kaiser, and Illia Polosukhin. 2017.
\newblock Attention is all you need.
\newblock In \emph{Advances in Neural Information Processing Systems}, pages
  6000--6010.

\end{thebibliography}

\appendix

\end{document}